\documentclass[11pt,a4paper]{article}
\usepackage{acl2015}
\usepackage{times}
\usepackage{latexsym}
\usepackage{url}
\usepackage{floatrow}
\newfloatcommand{capbtabbox}{table}[][\FBwidth]
\usepackage{xcolor}
\usepackage{amsmath, graphicx, subfigure, amsfonts, amssymb}
\usepackage{algorithmicx}
\usepackage{algorithm}

\usepackage[noend]{algpseudocode}
\usepackage{caption}
\usepackage{breqn}
\usepackage{multirow, array} %table utils
\usepackage{tikz-dependency}

 \usepackage[scaled=0.92]{helvet}
 \usepackage[linewidth=1pt]{mdframed}

\usepackage{enumitem}
\setlist{nolistsep,leftmargin=*} 

 \usepackage[disable]{todonotes} % notes not showed
 \newcommand{\comment}[1]{}  %comment not showed

\newcommand*{\imp}{\textcolor{black}} %SHOW updates

\newcommand\blfootnote[1]{%
  \begingroup
  \renewcommand\thefootnote{}\footnote{#1}%
  \addtocounter{footnote}{-1}%
  \endgroup
}

\NewEnviron{elaboration}{
\par
\begin{tikzpicture}
\node[rectangle,minimum width=\textwidth] (m) {\begin{minipage}{0.98\textwidth}\BODY\end{minipage}};
\draw[dashed] (m.south west) rectangle (m.north east);
\end{tikzpicture}
}

\title{Language Understanding for Text-based Games using Deep Reinforcement Learning}
 
\author{Karthik Narasimhan$^*$\\
      CSAIL, MIT\\
      {\tt karthikn@csail.mit.edu}
    \And
     Tejas D Kulkarni$^*$\\
      CSAIL, BCS, MIT\\
      {\tt tejask@mit.edu}
     \And
  Regina Barzilay\\
      CSAIL, MIT\\
      {\tt regina@csail.mit.edu}
}

\date{}

\begin{document}
\maketitle
\begin{abstract}
In this paper, we consider the task of learning control policies for text-based games.  In these games, all interactions in the virtual world are through text and the underlying state is not observed. The resulting language barrier makes such environments challenging for automatic game players. We employ a deep reinforcement learning framework to jointly learn state representations and action policies using game rewards as feedback. This framework enables us to map text descriptions into vector representations that capture the semantics of the game states.
\blfootnote{$^*$Both authors contributed equally to this work.}
 We evaluate our approach on two game worlds, comparing against \imp{baselines using bag-of-words and bag-of-bigrams for state representations.} 
 \imp{Our algorithm outperforms the baselines on both worlds demonstrating the importance of learning expressive representations.}
 \footnote{Code is available at \url{http://people.csail.mit.edu/karthikn/mud-play}.}
 
\end{abstract}
\vspace{0.2mm}
\section{Introduction}

In this paper, we address the task of learning control policies for text-based strategy games.  These games, predecessors to modern graphical ones, still enjoy a large following worldwide.\footnote{\url{http://mudstats.com/}} They often involve complex worlds with rich interactions and elaborate textual descriptions of the underlying states (see Figure~\ref{mud-example}). Players read descriptions of the current world state and respond with natural language commands to take actions. Since the underlying state is not directly observable, the player has to understand the text in order to act, making it challenging for existing AI programs
%~\cite{brian2009ai,huang2011skynet} 
to play these games~\cite{depristo2001being}.

\begin{figure}[t]
\begin{mdframed}

%  \noindent\fbox{%
\begin{elaboration}
  \parbox{0.98\textwidth}{
\emph{State 1: The old bridge} \\
You are standing very close to the bridge's eastern foundation. If you go east you will be back on solid ground ...
The bridge sways in the wind. 
}
\end{elaboration}

%\vspace{-1mm}
\begin{flushleft}
\emph{Command:} \textbf{Go east}
\end{flushleft}
%\vspace{2mm}

\begin{elaboration}
%  \noindent\fbox{%
  \noindent\parbox{0.99\textwidth}{
\emph{State 2: Ruined gatehouse} \\ 
The old gatehouse is near collapse. Part of its northern wall has
already fallen down ... 
%, together with parts of the fortifications in
%that direction.  \\
%... 
%Heavy stone pillars hold up sections of ceiling, but
%elsewhere the flagstones are exposed to open sky. Part of a heavy
%portcullis, formerly blocking off the inner castle from attack, is
%sprawled over the ground together with most of its frame.

East of the gatehouse leads out to a small open area surrounded by
the remains of the castle.  There is also a standing archway
offering passage to a path along the old southern inner wall. \\
Exits: Standing archway, castle corner, Bridge over the abyss
}
\end{elaboration}
\end{mdframed}

\caption{Sample gameplay from a Fantasy World. The player with the quest of finding a secret tomb, is currently located on an \emph{old bridge}. She then chooses an action to \emph{go east} that brings her to a \emph{ruined gatehouse} (State 2).
}
\label{mud-example}
 \end{figure}
 
In designing an autonomous game player, we have considerable latitude when selecting an adequate state representation to use. The simplest method is to use a bag-of-words representation derived from the text description. However, this scheme disregards the ordering of words and the finer nuances of meaning that evolve from composing words into sentences and paragraphs. For instance, in State 2 in Figure~\ref{mud-example}, the agent has to understand that going \emph{east} will lead it to the castle whereas moving \emph{south} will take it to the standing archway.
An alternative approach is to convert text descriptions to pre-specified representations using annotated training data, commonly used in language grounding tasks~\cite{matuszek2013learning,kushman2014learning}. 

In contrast, our goal is to learn useful representations in conjunction with control policies. We adopt a reinforcement learning framework and formulate game sequences as Markov Decision Processes. An agent playing the game aims to maximize rewards that it obtains from the game engine upon the occurrence of certain events. The agent learns a policy in the form of an action-value function $Q(s,a)$ which denotes the long-term merit of an action~$a$ in state~$s$. 

%We learn to play text-based games using a \emph{deep reinforcement learning} framework, which is trained based on game feedback. 

The action-value function is parametrized using a deep recurrent neural network, trained using the game feedback. The network contains two modules. The first one converts textual descriptions into vector representations that act as proxies for states. This component is implemented using Long Short-Term Memory (LSTM) networks ~\cite{hochreiter1997long}. The second module of the network scores the actions given the vector representation computed by the first.

We evaluate our model using two Multi-User Dungeon (MUD) games~\cite{curtis1992mudding,amir2002adventure}. \imp{The first game is designed to provide a controlled setup for the task, while the second is a publicly available one and contains human generated text descriptions with significant language variability.} We compare our algorithm against baselines of a random player and models that use bag-of-words or bag-of-bigrams representations for a state. We demonstrate that our model LSTM-DQN significantly outperforms the baselines in terms of number of completed quests and \imp{accumulated rewards}. For instance, on a fantasy MUD game, our model learns to complete 96\% of the quests, while the bag-of-words model and a random baseline solve only 82\%  and 5\% of the quests, respectively.  Moreover, we show that the acquired representation can be reused across games, speeding up learning and leading to faster convergence of Q-values. 

% add note for : LSTM not over episode states?
\section{Related Work}
%\imp{The area of language grounding has seen considerable advancement recently}
Learning control policies from text is gaining increasing interest in the NLP community. Example applications include interpreting help documentation for software~\cite{branavan2010reading}, navigating with directions~\cite{vogel2010learning,kollar2010toward,artzi2013weakly,matuszek2013learning,Andreas15Instructions} and playing computer games~\cite{eisenstein-EtAl:2009:EMNLP,branavan2011learning}.

 Games provide a rich domain for grounded language analysis.
% Access to an automatically computed reward, often explicit state and action representation, coupled with related textual resources, provides a powerful source of supervision for language interpretation. 
Prior work has assumed perfect knowledge of the underlying state of the game to learn policies.
 \newcite{DBLP:conf/aiide/GorniakR05} developed a game character that can be controlled by spoken instructions adaptable  to the game situation. The grounding of commands to actions is learned from a transcript manually annotated with actions and state attributes. \newcite{eisenstein-EtAl:2009:EMNLP} learn game rules by analyzing a collection of game-related documents and precompiled traces of the game. In contrast to the above work, our model combines text interpretation and strategy learning in a single framework. As a result, textual analysis is guided by the received control feedback, and the learned strategy directly builds on the text interpretation.

Our work closely relates to an automatic game player that utilizes text manuals to learn strategies for Civilization~\cite{branavan2011learning}. Similar to our approach, text analysis and control strategies are 
learned jointly using feedback provided by the game simulation. In their setup, states are fully observable, and the model learns a strategy by combining state/action features and features extracted from text.   However, in our application, the state representation is not provided, but has to be inferred from a textual description.  Therefore, it is not sufficient to extract features from text to supplement a simulation-based player. 
%\todo[]{check this wording}
%Instead, we have to learn representation directly from text.  

Another related line of work consists of automatic video game players that infer state representations directly from raw pixels~\cite{koutnik2013evolving,mnih2015dqn}.  For instance, \newcite{mnih2015dqn} learn control strategies using convolutional neural networks, trained with a variant of Q-learning~\cite{watkins1992q}. While both approaches use deep reinforcement learning for training, our work has important differences. In order to handle the sequential nature of text, we use Long Short-Term Memory networks to automatically learn useful representations \imp{for arbitrary text descriptions.} Additionally, we show that decomposing the network into a representation layer and an action selector is useful for transferring the learnt representations to new game scenarios.

% [[\imp{The following sentence is not true. Think of Montezuma's revenge and many other Atari games}]] Second, in our case, the agent only receives delayed feedback on completion of quests instead of frequently added points to game score. 
% We also add a priority sampling trick that helps speed up learning.

\section{Background}
%\todo[]{move to intro?}
%Multi-User Dungeon (MUD) games are typically real-time text-based virtual worlds supporting multiple players and interactions between them. Players read textual descriptions of rooms, objects, characters and actions, and respond with their own natural language commands to take actions. In spite of having less sensory input than graphical games, MUD games often involve complex worlds with rich interactions and varying textual descriptions of the underlying state(s). Games typically revolve around character building by learning skills and completing quests, while also interacting with non-player characters (NPCs) and other players. Figure \ref{mud-example} provides an example of the gameplay of a typical MUD. 

\paragraph{Game Representation} 
We represent a game by the tuple $\langle H, A, T, R, \Psi \rangle$, where $H$ is the set of all possible game states, $A = \{(a,o)\}$ is the set of all commands (action-object pairs), $T(h' \mid h, a, o)$ is the stochastic transition function between states and $R(h, a, o)$ is the reward function. The game state $H$ is \emph{hidden} from the player, who only receives a varying textual description, produced by a stochastic function $\Psi : H \to S$. Specifically, the underlying state $h$ in the game engine keeps track of attributes such as the player's location, her health points, time of day, etc. The function $\Psi$ (also part of the game framework) then converts this state into a textual description of the location the player is at or a message indicating low health. We do not assume access to either $H$ or $\Psi$ for our agent during both training and testing phases of our experiments. We denote the space of all possible text descriptions $s$ to be $S$. Rewards are generated using $R$ and are only given to the player upon completion of in-game quests. 

\paragraph{Q-Learning}
Reinforcement Learning is a commonly used framework for learning control policies in game environments~\cite{silver2007reinforcement,amato2010high,branavan2011nonlinear,szita2012reinforcement}. The game environment can be formulated as a sequence of state transitions $(s,a,r,s')$ of a Markov Decision Process (MDP). The agent takes an action $a$ in state $s$ by consulting a state-action value function $Q(s,a)$, which is a measure of the action's expected long-term reward. Q-Learning~\cite{watkins1992q} is a model-free technique which is used to learn an optimal $Q(s,a)$ for the agent. Starting from a random Q-function, the agent continuously updates its Q-values by playing the game and obtaining rewards. The iterative updates are derived from the Bellman equation~\cite{sutton1998introduction}:
\begin{dmath}
	{Q_{i+1}(s,a) = \mathrm{E}[r + \gamma \max_{a'} Q_i(s',a') \mid s, a]}
\label{eq:bellman-update}
\end{dmath}
where  $\gamma$ is a discount factor for future rewards and  the expectation is over all game transitions that involved the agent taking action $a$ in state $s$.

Using these evolving Q-values, the agent chooses the action with the highest $Q(s,a)$ to maximize its expected future rewards. In practice, the trade-off between exploration and exploitation can be achieved following an $\epsilon$-greedy policy~\cite{sutton1998introduction}, where the agent performs a random action with probability $\epsilon$.

%since the Q-values need to be learnt as the agent is playing, there is a trade-off to be made between choosing the best action according to the current Q-value and exploring new actions (to potentially uncover a better action). This can be achieved by following an $\epsilon$-greedy policy,  in which the agent chooses a random action with probability $\epsilon$ and the best action according to its Q-value function with probability $1-\epsilon$. 

\paragraph{Deep Q-Network}
In large games, it is often impractical to maintain the Q-value for all possible state-action pairs. One solution to this problem is to approximate $Q(s,a)$ using a parametrized function $Q(s,a; \theta)$, which can generalize over states and actions by considering higher-level attributes~\cite{sutton1998introduction,branavan2011learning}. However, creating a good parametrization requires knowledge of the state and action spaces. One way to bypass this feature engineering is to use a Deep Q-Network (DQN)~\cite{mnih2015dqn}. The DQN approximates the Q-value function with a deep neural network to predict $Q(s,a)$ for all possible actions $a$ simultaneously given the current state $s$. The non-linear function layers of the DQN also enable it to learn better value functions than linear approximators.

\section{Learning Representations and Control Policies}
\begin{figure}[t]
\centering
\resizebox{1.05\columnwidth}{!}{%
\includegraphics[]{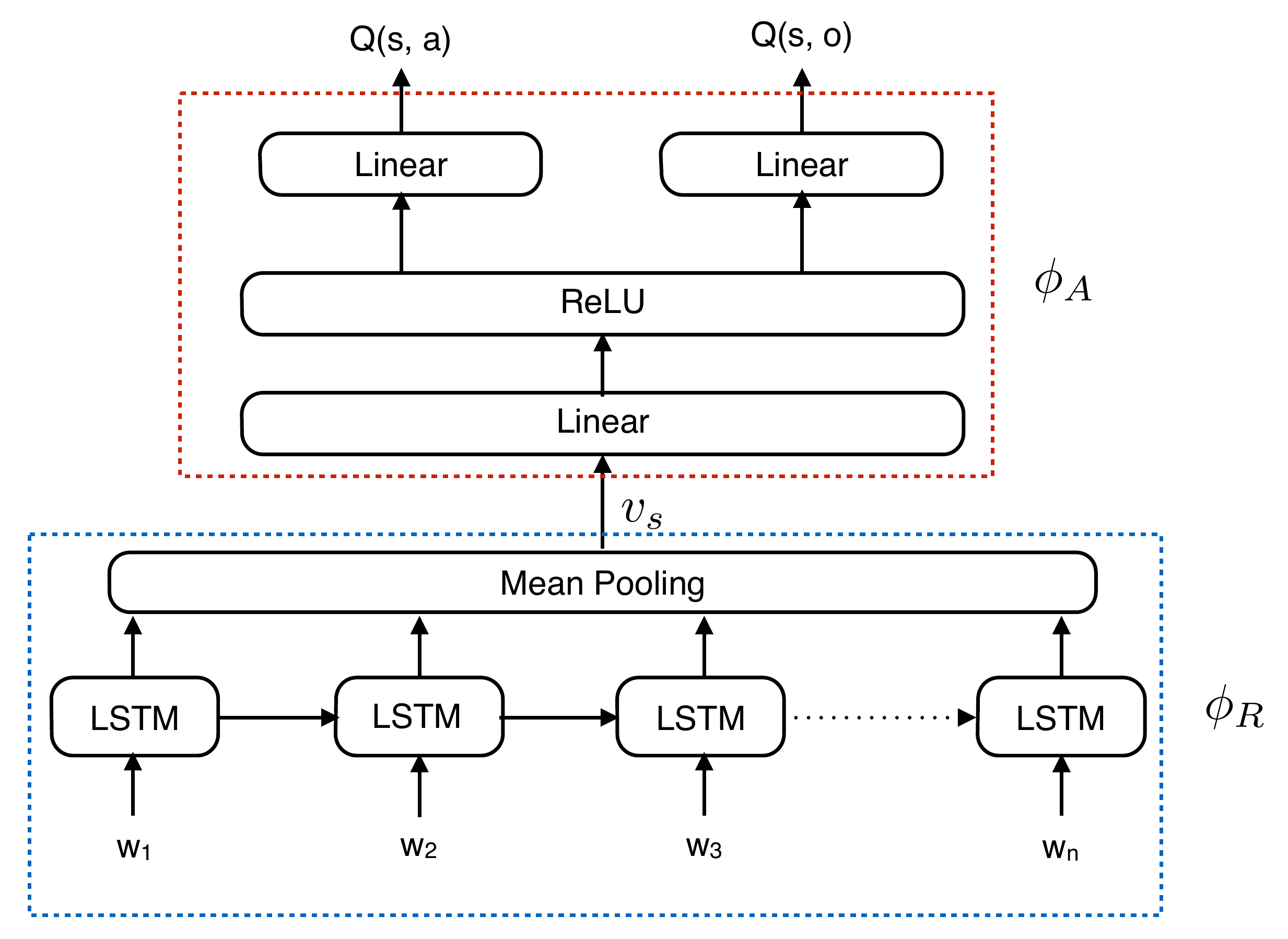}
}
\caption{Architecture of LSTM-DQN: The Representation Generator ($\phi_R$) (bottom) takes as input  a stream of words observed in state $s$ and produces a vector representation $v_s$, which is fed into the action scorer ($\phi_A$) (top) to produce scores for all actions and argument objects.}
\label{fig:lstm-dqn}
\end{figure}

%Textual MUDs are challenging due to the stochastic textual descriptions and the large state-action space.
In this section, we describe our model (DQN) and describe its use in learning good Q-value approximations for games with stochastic textual descriptions.
 We divide our model into two parts.  
The first module is a \emph{representation generator} that converts the textual description of the current state into a vector. This vector is then input into the second module which is an \emph{action scorer}. Figure~\ref{fig:lstm-dqn} shows the overall architecture of our model. We learn the parameters of both the representation generator and the action scorer jointly, using the in-game reward feedback.  
 
\paragraph{Representation Generator ($\phi_R$)}
The representation generator reads raw text displayed to the agent and converts it to a vector representation $v_s$. A bag-of-words (BOW) representation is not sufficient to capture higher-order structures of sentences and paragraphs. The need for a better semantic representation of the text is evident from the average performance of this representation in playing MUD-games (as we show in Section \ref{sec:results}).

In order to assimilate better representations, we utilize a Long Short-Term Memory network (LSTM)~\cite{hochreiter1997long} as a representation generator. LSTMs are recurrent neural networks with the ability to connect and recognize long-range patterns between  words in text. They are more robust than BOW to small variations in word usage and are able to capture underlying semantics of sentences to some extent. In recent work, LSTMs have been used successfully in NLP tasks such as machine translation~\cite{sutskever2014sequence} and sentiment analysis~\cite{tai2015improved} to compose vector representations of sentences from word-level embeddings~\cite{mikolov2013efficient,pennington2014glove}. In our setup, the LSTM network takes in word embeddings $w_k$ from the words in a description $s$ and produces output vectors $x_k$ at each step.

%An LSTM cell consists of an input gate $i$, a forget gate $f$ and a memory cell $c$. At each step~$k$ over the words\footnote{We convert all the words into vectors which are input to the LSTM cell.}  in a text description $s~=~\langle w_1, w_2, ... w_n \rangle$, the LSTM takes in the following vectors in $\mathbb{R}^d$: an input word $w_k$, the output from the previous step $h_{k-1}$ and a memory vector $c_{k-1}$, also from the previous step. At each step, the cell produces an output vector $x_k$, which is an accumulated representation over the preceding steps.

%\todo[inline]{add LSTM figure?}

%The transition equations for the LSTM can be summarized as:
%
%\begin{align}
%\label{eqn:eqlabel}
%\begin{split}
%	i_k &= \sigma (U^{(i)} w_k + V^{(i)} h_{k-1} + b^{(i)}), \\
%	f_k &= \sigma (U^{(f)} w_k + V^{(f)} h_{k-1} + b^{(f)}), \\
%	o_k &= \sigma (U^{(o)} w_k + V^{(o)} h_{k-1} + b^{(o)}) \\
%	z_k &= \tanh ( U^{(z)} w_k + V^{(z)} h_{k-1} + b^{(z)})	\\
%	c_k &= i_k \odot z_k + f_k \odot c_{k-1} \\
%	x_k &= o_k \odot \tanh(c_k)
%\end{split}
%\end{align}
%where $\sigma$ represents the sigmoid function and $\odot$ is elementwise multiplication.
 To get the final state representation $v_s$, we add a \emph{mean pooling} layer which computes the element-wise mean over the output vectors $x_k$.\footnote{We also experimented with considering just the output vector of the LSTM after processing the last word. Empirically, we find that mean pooling leads to faster learning, so we use it in all our experiments.}
 \begin{dmath}
 	v_s = \frac{1}{n} \sum_{k=1}^{n} x_k
 \end{dmath}

 \begin{algorithm*}
\caption{Training Procedure for DQN with prioritized sampling}
\label{alg:training}

\begin{algorithmic}[1]
\State Initialize experience memory $\mathcal{D}$
\State Initialize parameters of representation generator ($\phi_R$) and action scorer ($\phi_A$) randomly
\For {$ episode = 1,M $}
	\State Initialize game and get start state description $s_1$	
	\For {$ t = 1, T $}
		\State Convert $s_t$ (text) to representation $v_{s_t}$ using $\phi_R$
		\If {$ random() < \epsilon $}			
			\State Select a random action $a_t$
		\Else		
			\State Compute $Q(s_t, a)$ for all actions using $\phi_A(v_{s_t})$
			\State Select  $a_t = \text{argmax}~Q(s_t, a)$  
		\EndIf
		\State Execute action $a_t$ and observe reward $r_t$ and new state $s_{t+1}$
		\State Set priority $p_t = 1$ if $r_t > 0$, else $p_t = 0$
		\State Store transition $(s_t, a_t, r_t, s_{t+1}, p_t)$ in $\mathcal{D}$
		\State Sample random mini batch of transitions $(s_j, a_j, r_j, s_{j+1}, p_j)$ from $\mathcal{D}$, \par
			 \hskip\algorithmicindent with fraction $\rho$ having $p_j = 1$
		\State Set $y_j = \left\{ 
		\begin{array}{lcl}
 				       r_j &\text{if } s_{j+1} \text{ is terminal}\\
				       r_j + \gamma~\text{max}_{a'}~Q(s_{j+1}, a'; \theta) &\text{if } s_{j+1} \text{ is non-terminal}
	      \end{array}
		    \right. $
		\State Perform gradient descent step on the loss $\mathcal{L}(\theta) = (y_j - Q(s_j, a_j; \theta))^2$ 
	\EndFor
\EndFor

\end{algorithmic}
\end{algorithm*}

\paragraph{Action Scorer ($\phi_A$)}
 The action scorer module produces scores for the set of possible actions given the current state representation. We use a multi-layered neural network for this purpose (see Figure~\ref{fig:lstm-dqn}).  The input to this module is the vector from the representation generator, $v_s = \phi_R(s)$ and the outputs are scores for actions $a \in A$.  Scores for all actions are predicted simultaneously, which is computationally more efficient than scoring each state-action pair separately. Thus, by combining the representation generator and action scorer, we can obtain the approximation for the Q-function as $Q(s,a) \approx \phi_A (\phi_R(s)) [a] $.
 
 An additional complexity in playing MUD-games is that the actions taken by the player are multi-word natural language \emph{commands} such as \emph{eat apple} or \emph{go east}. Due to computational constraints, in this work we limit ourselves to consider commands to consist of one action (e.g. \emph{eat}) and one argument object (e.g. \emph{apple}). This assumption holds for the majority of the commands in our worlds, with the exception of one class of commands that require two arguments  (e.g. \emph{move red-root right}, \emph{move blue-root up}).
 We consider all possible actions and objects available in the game and predict both for each state using the same network (Figure \ref{fig:lstm-dqn}). We consider the Q-value of the entire command $(a,o)$ to be the average of the Q-values of the action $a$ and the object $o$. For the rest of this section, we only show equations for $Q(s, a)$ but similar ones hold for $Q(s, o)$.

\paragraph{Parameter Learning}
We learn the parameters $\theta_R$ of the representation generator and $\theta_A$ of the action scorer using stochastic gradient descent with \emph{RMSprop}~\cite{tieleman2012lecture}. The complete training procedure is shown in Algorithm \ref{alg:training}.
 In each iteration $i$, we update the parameters to reduce the discrepancy between the predicted value of the current state $Q(s_t, a_t; \theta_i)$ (where $\theta_i = [\theta_R;\theta_A]_i$) and the expected Q-value given the reward $r_t$ and the value of the next state $\text{max}_a~Q(s_{t+1}, a; \theta_{i-1})$. 
 
 We keep track of the agent's previous experiences in a \emph{memory} $\mathcal{D}$.\footnote{The memory is limited and rewritten in a first-in-first-out (FIFO) fashion.} Instead of performing updates to the Q-value using transitions from the current episode, we sample a random transition $(\hat{s}, \hat{a}, s', r)$  from $\mathcal{D}$. Updating the parameters in this way avoids issues due to strong correlation when using transitions of the same episode~\cite{mnih2015dqn}.
 Using the sampled transition and \eqref{eq:bellman-update}, we obtain the following loss function to minimize:
\begin{dmath}
	{\mathcal{L}_i(\theta_i) = \mathrm{E}_{\hat{s},\hat{a}}  [ (y_i - Q(\hat{s}, \hat{a} ; \theta_i))^2 ]}
\label{eq:loss}
\end{dmath}
where $ {y_i = \mathrm{E}_{\hat{s},\hat{a}}[r + \gamma \max_{a'} Q (s',a'; \theta_{i-1}) \mid \hat{s}, \hat{a}]}$ is the target Q-value with parameters $\theta_{i-1}$ fixed from the previous iteration.  

The updates on the parameters $\theta$ can be performed using the following gradient of $\mathcal{L}_i(\theta_i)$:
\begin{dmath*}
	{\nabla_{\theta_i} \mathcal{L}_i(\theta_i) = \mathrm{E}_{\hat{s},\hat{a}}  [ 2(y_i - Q(\hat{s}, \hat{a} ; \theta_i)) \nabla_{\theta_i} Q(\hat{s}, \hat{a} ; \theta_i) ]} 
\label{eq:loss-gradient}
\end{dmath*}
\vspace{-2mm}
For each epoch of training, the agent plays several episodes of the game, which is restarted after every terminal state.

\paragraph{Mini-batch Sampling}
In practice, online updates to the parameters $\theta$ are performed over a mini batch of state transitions, instead of a single transition. This increases the number of experiences used per step and is also more efficient due to optimized matrix operations. 
 
 The simplest method to create these mini-batches from the experience memory $\mathcal{D}$ is to sample uniformly at random. However, certain experiences are more valuable than others for the agent to learn from. For instance, rare transitions that provide positive rewards can be used more often to learn optimal Q-values faster. In our experiments, we consider such positive-reward transitions to have higher \emph{priority} and keep track of them in $\mathcal{D}$. We use \emph{prioritized sampling} (inspired by \newcite{moore1993prioritized}) to sample a fraction $\rho$ of transitions from the higher priority pool and a fraction $1-\rho$ from the rest.

\section{Experimental Setup}
\label{sec:experiments} 
\paragraph{Game Environment}
For our game environment, we modify Evennia,\footnote{\url{http://www.evennia.com/}} an open-source library for building online textual MUD games. \text{Evennia} is a Python-based framework that allows one to easily create new games by writing a batch file describing the environment
with details of rooms, objects and actions. The game engine keeps track of the game state internally, presenting textual 
descriptions to the player and receiving text commands from the player. We conduct experiments on two worlds - a smaller \emph{Home world} we created ourselves, and a larger, more complex \text{\emph{Fantasy world}} created by Evennia's developers. 
The motivation behind Home world is to abstract away high-level planning and focus on the language understanding requirements of the game.

\begin{table}[t]
\centering
\resizebox{\textwidth}{!}{%
\begin{tabular}{| c | c | c |} \hline
\textbf{Stats} & \textbf{Home World} & \textbf{Fantasy World} \\ \hline
Vocabulary size & 84 & 1340 \\ 
Avg. words / description  & 10.5 & 65.21 \\
Max descriptions / room & 3 & 100 \\
\# diff. quest descriptions & 12 & - \\ 
State transitions & Deterministic & Stochastic \\ 
\# states (underlying) & 16 &   $\geq$ 56 \\ 
Branching factor & & \\ 
(\# commands / state) & 40 & 222 \\ \hline
\end{tabular}
}
\caption{Various statistics of the two game worlds}
\label{table:game-stats}
\end{table}

Table~\ref{table:game-stats} provides statistics of the game worlds. \imp{We observe that the Fantasy world is moderately sized with a vocabulary of 1340 words and up to 100 different descriptions for a room. These descriptions were created manually by the game developers.  These diverse, engaging descriptions are designed to make it interesting and exciting for human players.  Several rooms have many alternative descriptions, invoked randomly on each visit by the player.}

\imp{Comparatively, the Home world is smaller: it has a very restricted vocabulary of 84 words and the room descriptions are relatively structured. However, both the room descriptions (which are also varied and randomly provided to the agent) and the quest descriptions were adversarially created with negation and conjunction of facts to force an agent to actually understand the state in order to play well. Therefore, this domain provides an interesting challenge for language understanding. }

In both worlds, the agent receives a positive reward on completing a quest, and negative rewards for getting into bad situations like falling off a bridge, or losing a battle. We also add small deterministic negative rewards for each non-terminating step. This incentivizes the agent to learn policies that solve quests in fewer steps. The supplementary material has details on the reward structure.
% and Table~\ref{table:reward-structure} shows the reward structure for the worlds.

\paragraph{Home World}
We created \emph{Home world} to mimic the environment of a typical house.\footnote{An illustration is provided in the supplementary material.} 
%Figure~\ref{fig:home-world} is an illustration of this world.
 The world consists of four rooms - a living room,
a bedroom, a kitchen and a garden with connecting pathways. Every room is reachable from every other room. Each room contains a representative object that the agent can interact with. For instance, the
kitchen has an \emph{apple} that the player can \emph{eat}. Transitions between the rooms are deterministic. At the start of each game episode, the player is placed in a random room and provided with a randomly selected quest. The text provided to the player contains both the description of her current state and that of the quest. Thus, the player can begin in one of 16 different states (4 rooms $\times$ 4 quests), which adds to the world's complexity.

 An example of a quest given to the player in text is \emph{Not you are sleepy now but you are hungry now}. To complete this quest and obtain a reward, the player has to navigate through the house to reach the kitchen and eat the apple (i.e type in the command \emph{eat apple}). More importantly, the player should interpret that the quest does not require her to take a nap in the bedroom. We created such misguiding quests to make it hard for agents to succeed without having an adequate level of language understanding. 

%\begin{figure}[t]
%\centering
%\resizebox{0.9\columnwidth}{!}{%
%\includegraphics[]{fig/game-world1}
%}
%\caption{Rooms and objects in the Home World with connecting pathways.}
%\label{fig:home-world}
%\end{figure}

\paragraph{Fantasy World}
%We also run experiments on a bigger \emph{Fantasy world} created by developers of Evennia in our experiments. 
%\imp{Figure~\ref{fig:tutorial-world} provides a graphical representation of the game. }
The Fantasy world is considerably more complex and involves quests such as navigating through a broken bridge or finding the secret tomb of an ancient hero. This game also has stochastic transitions in addition to varying state descriptions provided to the player. For instance, there is a possibility of the player falling from the bridge if she lingers too long on it. 
\todo[inline]{add more desc}
% or being attacked (and defeated) by the guardian patrolling the tombs. 

%Again, we setup the reward structure to give positive rewards to the agent on completing quests within the game such as crossing the bridge safely, defeating the guardian or reaching the secret tomb. The agent receives negative rewards on failures like falling off the bridge (a 5\% chance in the game environment) or losing to the guardian in combat. This is in addition to the per-step deterministic negative rewards, similar to those in the previous world. See Table \ref{table:reward-structure} for details.

%We setup the reward structure to give the agent a reward of +1 if it successfully enters the correct tomb. Similar to the \emph{Home world}, we have rewards of -0.01 for every step leading to a non-goal state and -0.1 for invalid commands. We also add rewards of -0.5 if the agent falls off the bridge (there is a 5\% chance of this for every step the agent takes on the bridge) and a reward of -0.5 if defeated by the guardian. We find that a single positive reward at the end of the game is not sufficient to learn from. The agent often learns a policy to avoid the broken bridge (since that has the potential for more negative reward). Hence, we also added in a reward of +0.5 for crossing the bridge successfully (only once per episode). We ran each episode for a maximum of XXX steps.

Due to the large command space in this game,\footnote{We consider 222 possible command combinations of  6 actions and 37 object arguments.} we make use of cues provided by the game itself to narrow down the set of possible objects to consider in each state. For instance, in the MUD example in Figure~1, the game provides a list of possible exits. If the game does not provide such clues for the current state, we consider all objects in the game.

%The puzzle in the game provides a unique challenge since the player is required to remember the description of an image on the 
%obelisk and use that to select the right tomb to enter, failing which he is trapped into the dark cell. 

\paragraph{Evaluation}
We use two metrics for measuring an agent's performance: (1) the cumulative reward obtained per episode averaged over the episodes and (2) the fraction of quests completed by the agent. 
%Similar to \newcite{mnih2015dqn}, we also track the average max Q-value ($\max_a Q(s,a)$) for a held-out validation set\footnote{The validation set is created by aggregating states visited by a random agent over several episodes.} of states which provides insight into the evolution of the agent's policy. 
The evaluation procedure is as follows. In each epoch, we first train the agent on $M$ episodes of $T$ steps each. At the end of this training, we have a testing phase of running $M$ episodes of the game for $T$ steps. We use $M=50, T=20$ for the Home world and $M=20, T=250$ for the Fantasy world. For all evaluation episodes, we run the agent following an $\epsilon$-greedy policy with $\epsilon=0.05$, which makes the agent choose the best action according to its Q-values 95\% of the time. We report the agent's performance at each epoch.

\paragraph{Baselines}
We compare our LSTM-DQN model with three baselines. The first is a \emph{Random} agent that chooses both actions and objects uniformly at random from all available choices.\footnote{In the case of the Fantasy world, the object choices are narrowed down using game clues as described earlier.}  \imp{The other two are BOW-DQN and BI-DQN, which use a bag-of-words and a bag-of-bigrams representation of the text, respectively, as input to the DQN action scorer. These baselines serve to illustrate the importance of having a good representation layer for the task. }

\paragraph{Settings}
For our DQN models, we used $\mathcal{D} = 100000, \gamma = 0.5$. We use a learning rate of 0.0005 for RMSprop. We anneal the $\epsilon$ for $\epsilon$-greedy from 1 to 0.2 over 100000 transitions. A mini-batch gradient update is performed every 4 steps of the gameplay. We roll out the LSTM (over words) for a maximum of 30 steps on the Home world and for 100 steps on the Fantasy world. For the prioritized sampling, we used $\rho = 0.25$ for both worlds. We employed a mini-batch size of 64 and word embedding size $d=20$ in all experiments.

\begin{figure*}[!ht]
\RawFloats
\minipage{0.32\textwidth}
  \includegraphics[width=\linewidth]{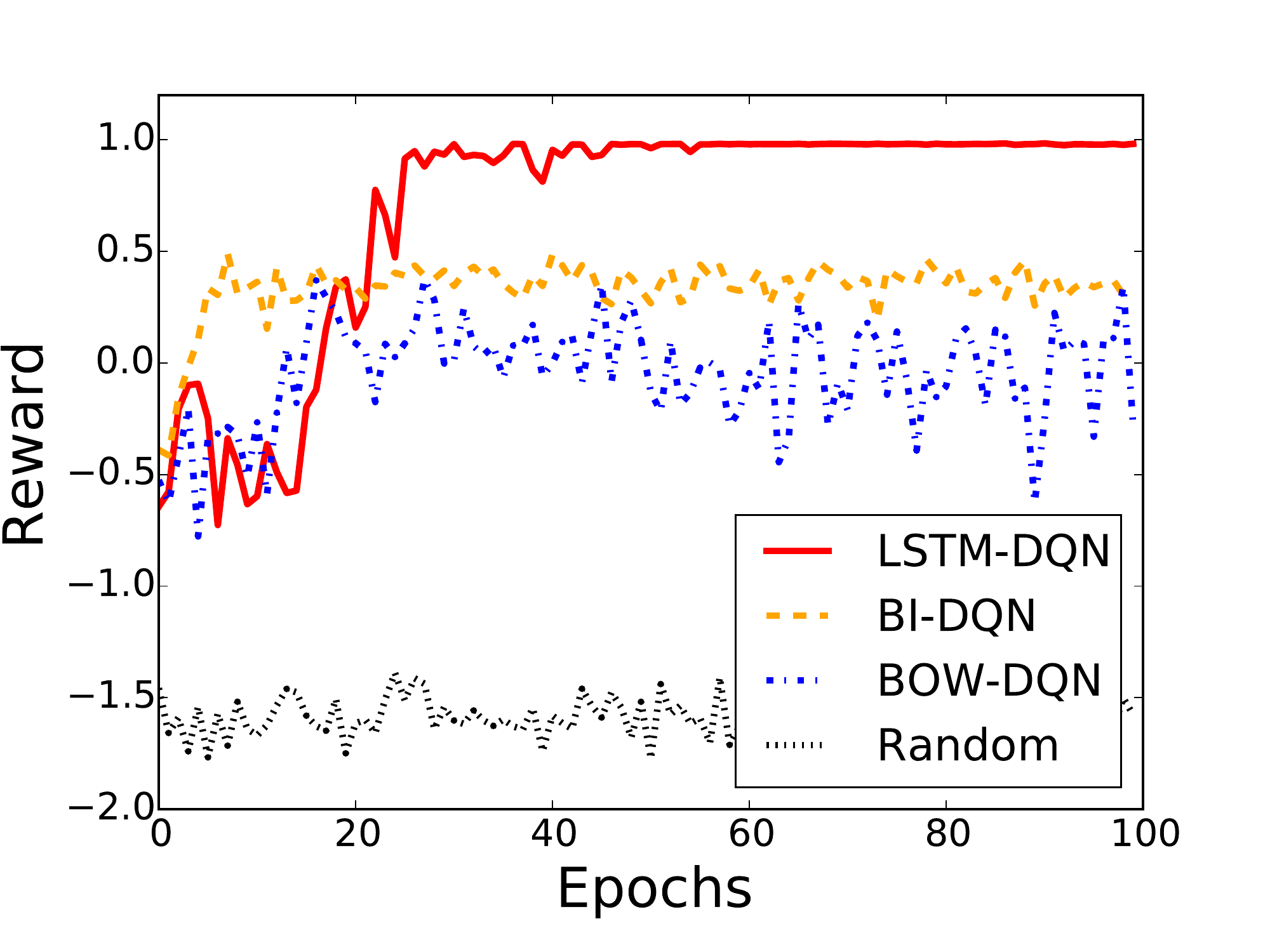}
        \caption*{Reward (Home)} 
\endminipage\hfill
\minipage{0.32\textwidth}%
  \includegraphics[width=\linewidth]{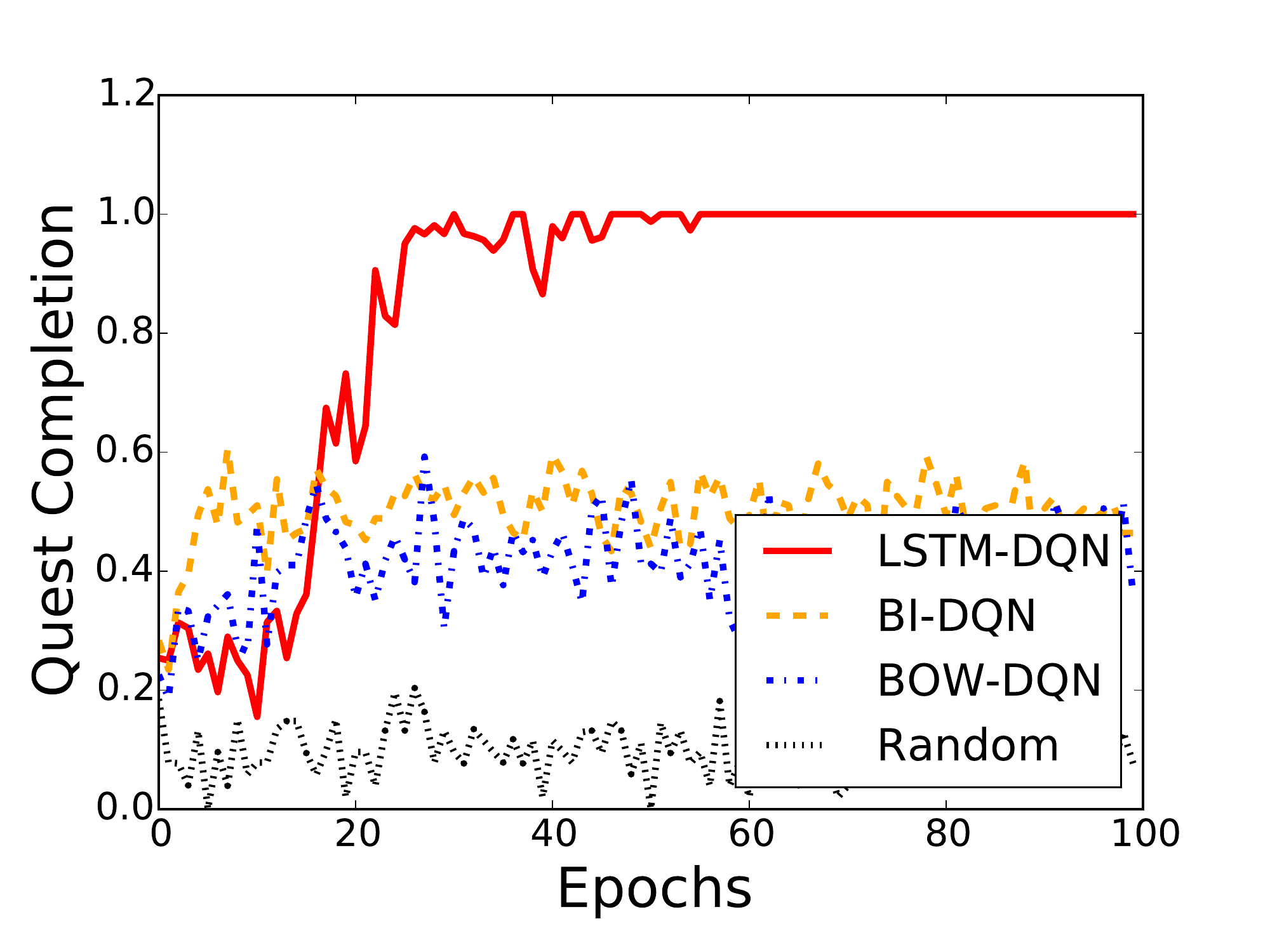}
        \caption*{Quest completion (Home)} 
\endminipage\hfill	
{\color{black}\vrule}\hfill	
\minipage{0.32\textwidth}
  \includegraphics[width=\linewidth]{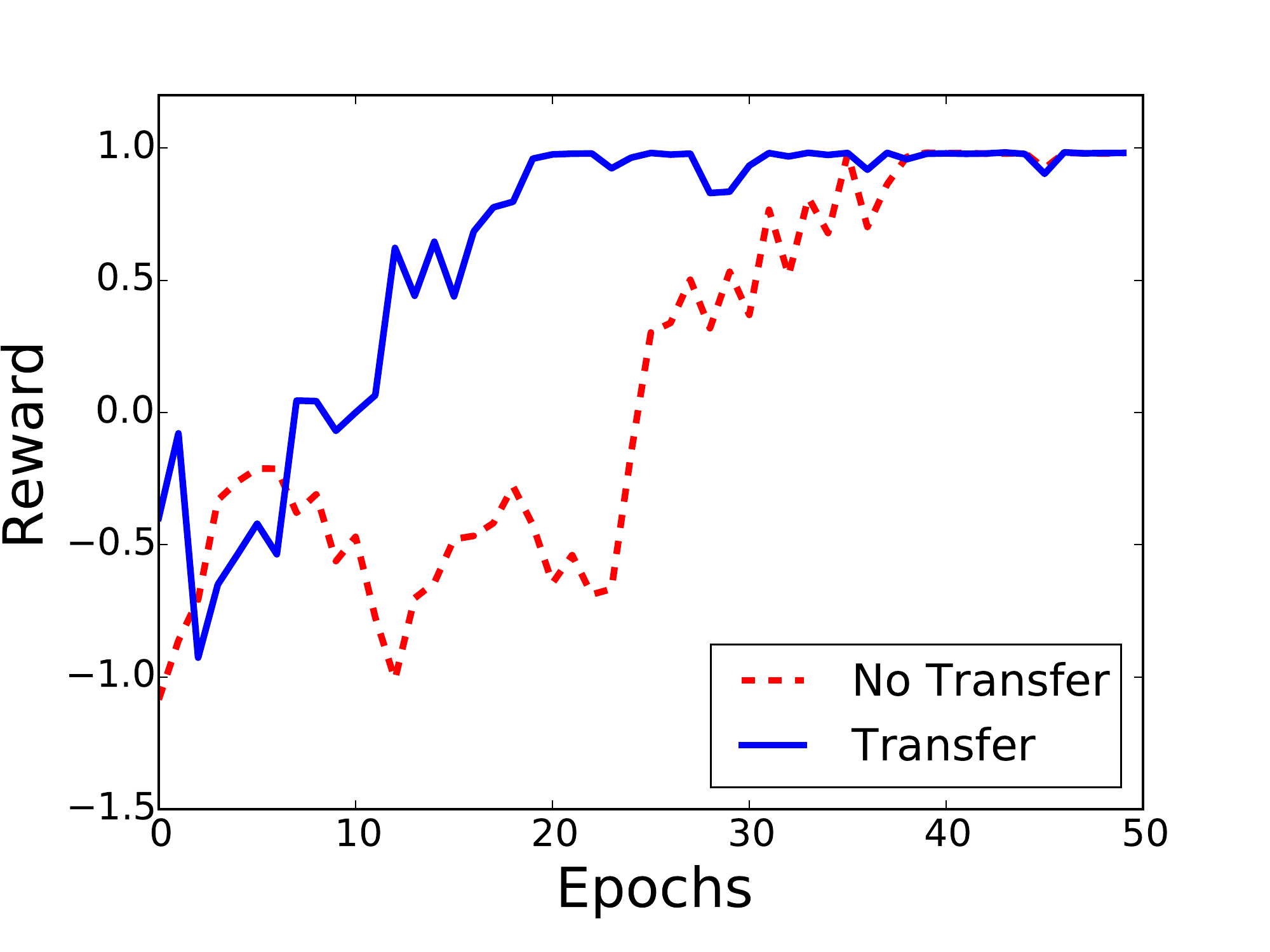}
    \caption*{Transfer Learning (Home)} 
    \label{fig:home-transfer-avgR}
\endminipage

\minipage{0.32\textwidth}
  \includegraphics[width=\linewidth]{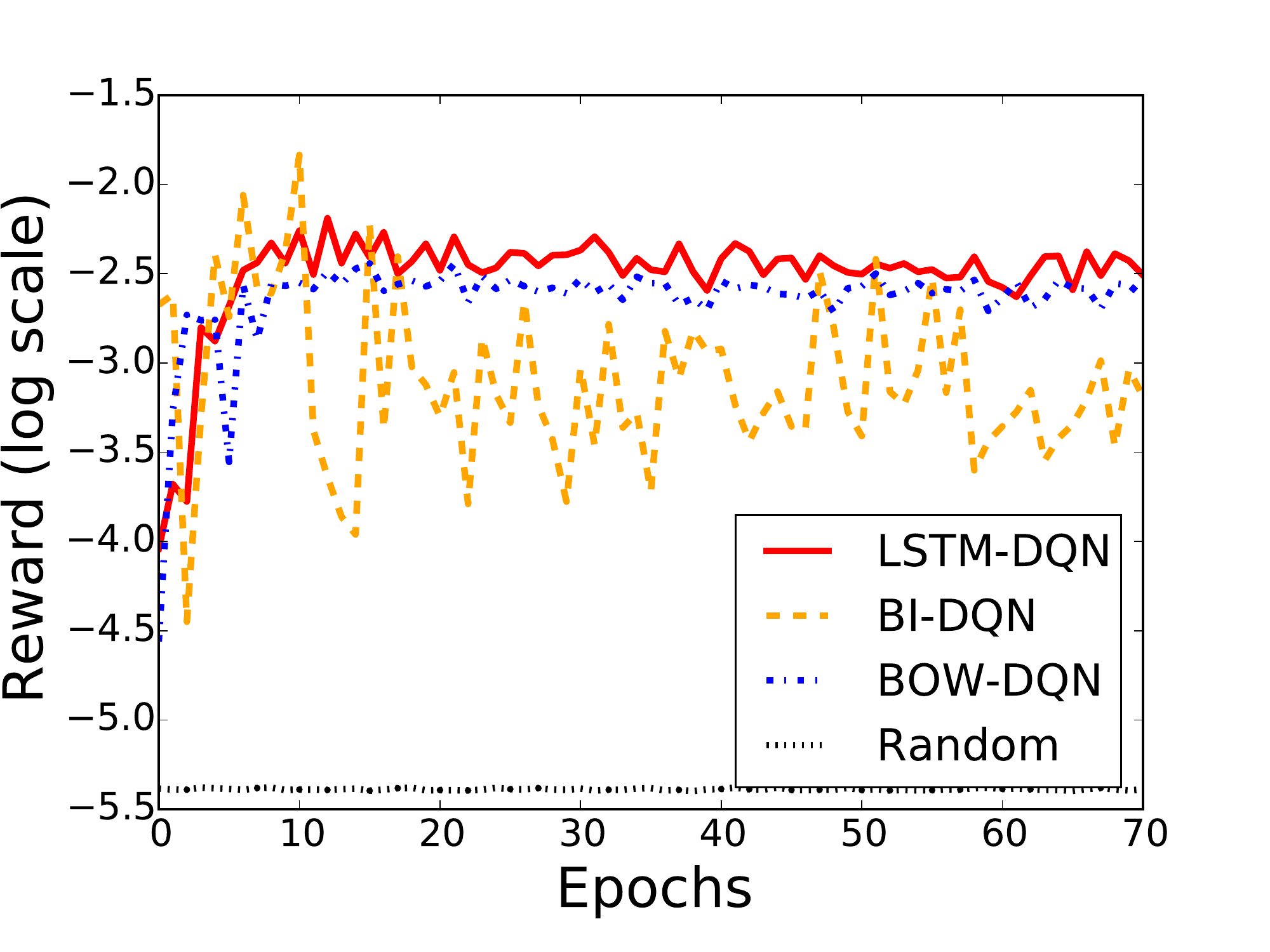}
    \caption*{Reward (Fantasy)} 
\endminipage\hfill
\minipage{0.32\textwidth}%
  \includegraphics[width=\linewidth]{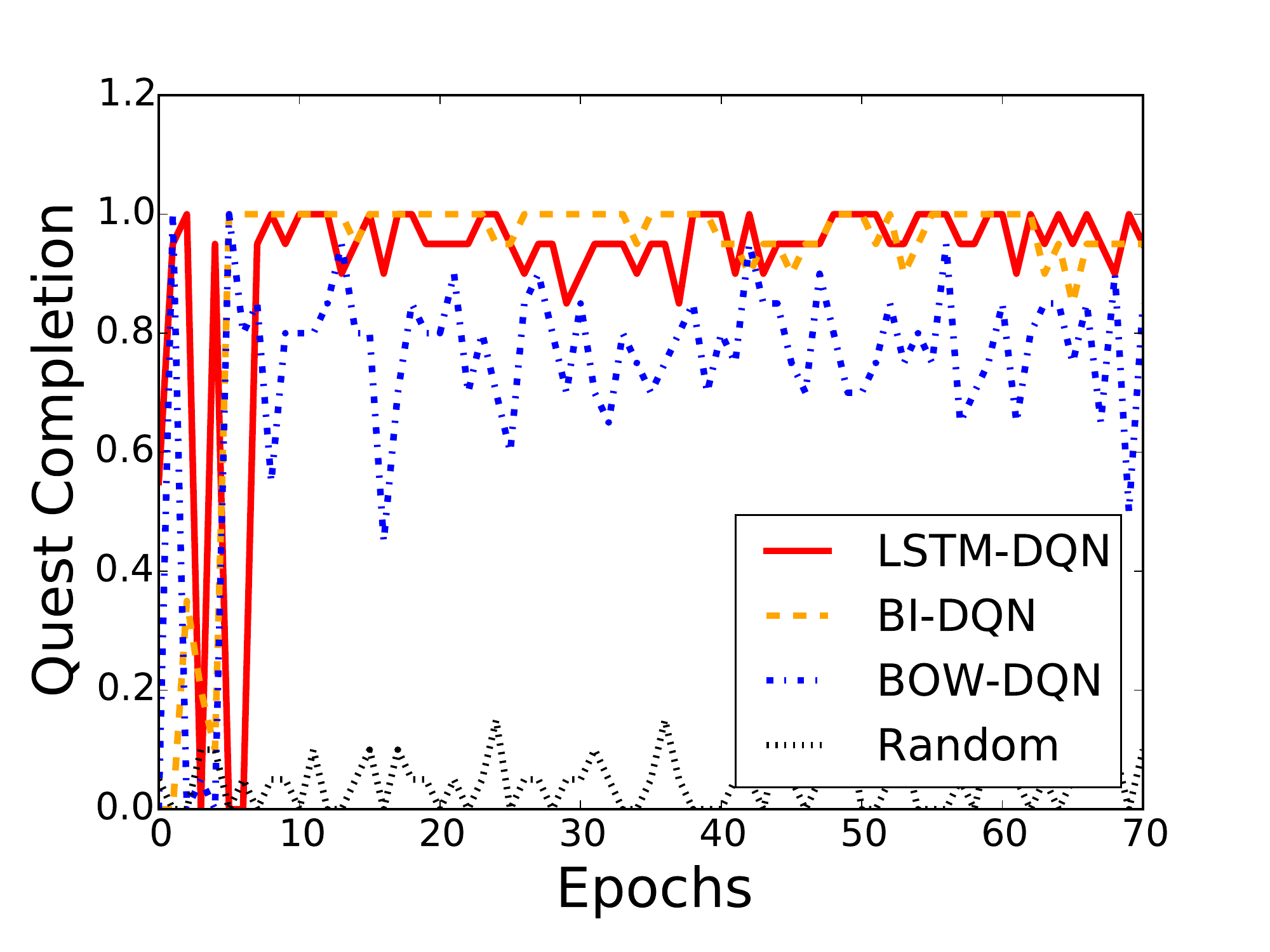}
      \caption*{Quest completion (Fantasy)} 
\endminipage\hfill
{\color{black}\vrule}\hfill	
\minipage{0.32\textwidth}
  \includegraphics[width=\linewidth]{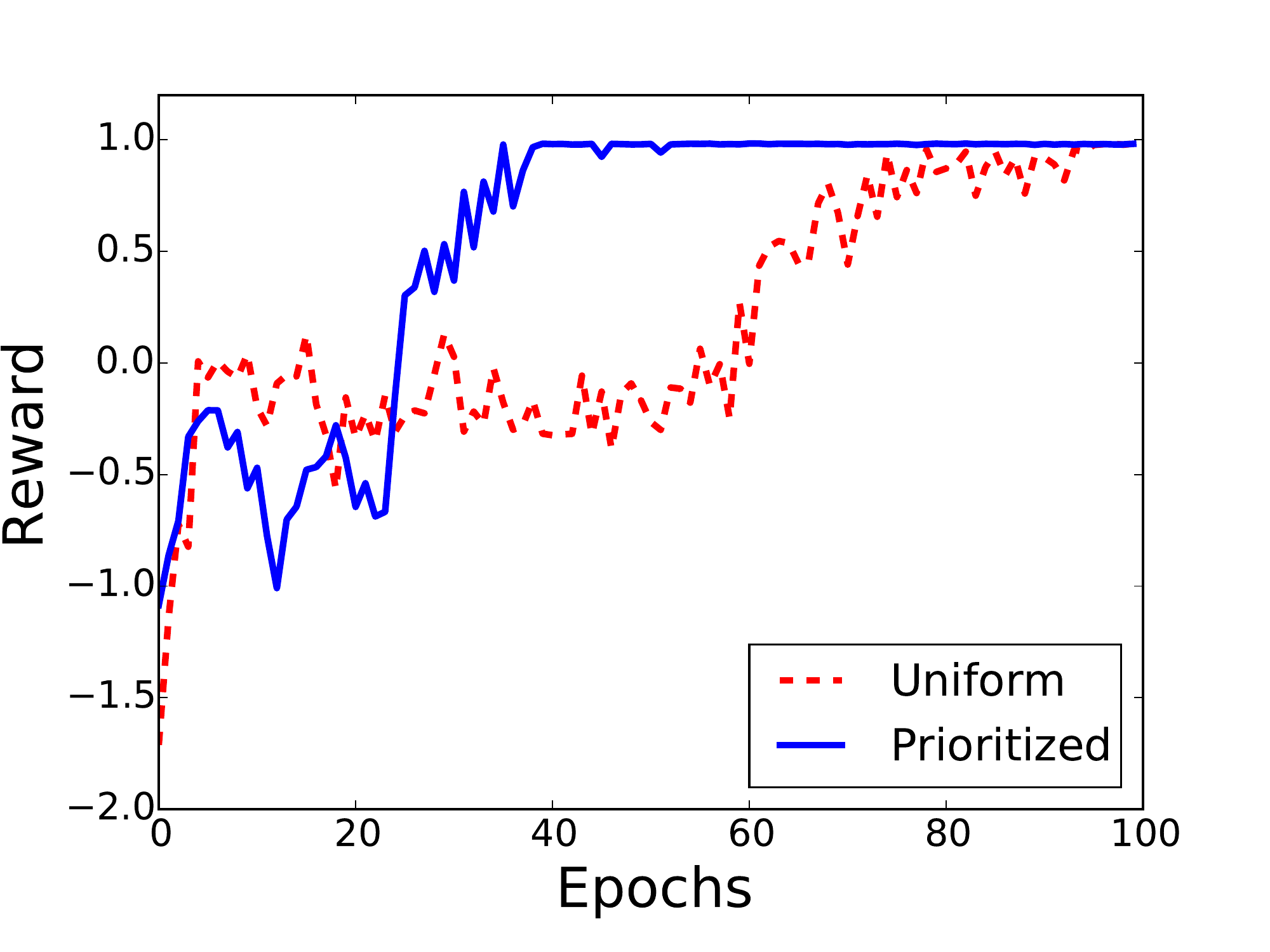}
  \caption*{Prioritized Sampling (Home)} 
  \label{fig:home-priority-avgR}
\endminipage
\caption{\textbf{Left:} Graphs showing the evolution of average reward  and quest completion rate for BOW-DQN, LSTM-DQN and a Random baseline on the Home world (\textbf{top}) and Fantasy world (\textbf{bottom}). Note that the reward is shown in log scale for the Fantasy world. \textbf{Right:} Graphs showing effects of transfer learning and prioritized sampling on the Home world. }
	\label{fig:results}
\end{figure*}

\section{Results}
\label{sec:results}

%\subsection{Main Results}
\paragraph{Home World}
Figure \ref{fig:results} illustrates the performance of LSTM-DQN compared to the baselines. We can observe that the Random baseline performs quite poorly, completing only around 10\% of quests on average\footnote{Averaged over the last 10 epochs.} obtaining a low reward of around $-1.58$. The BOW-DQN model performs significantly better and is able to complete around 46\% of the quests, with an average reward of $0.20$. The improvement in reward is due to both greater quest success rate and a lower rate of issuing invalid commands (e.g. \emph{eat apple} would be invalid in the bedroom since there is no apple). We notice that both the reward and quest completion graphs of this model are volatile. This is because the model fails to pick out differences between quests like \emph{Not you are hungry now but you are sleepy now}  and \emph{Not you are sleepy now but you are hungry now}. \imp{The BI-DQN model suffers from the same issue although it performs slightly better than BOW-DQN by completing 48\% of quests.} In contrast, the LSTM-DQN model does not suffer from this issue and is able to complete 100\% of the quests after around 50 epochs of training, achieving close to the optimal reward possible.\footnote{Note that since each step incurs a penalty of $-0.01$, the best reward (on average) a player can get is around $0.98$.} \imp{This demonstrates that having an expressive representation for text is crucial to understanding the game states and choosing intelligent actions.}

\begin{figure}
  \includegraphics[width=0.95\linewidth]{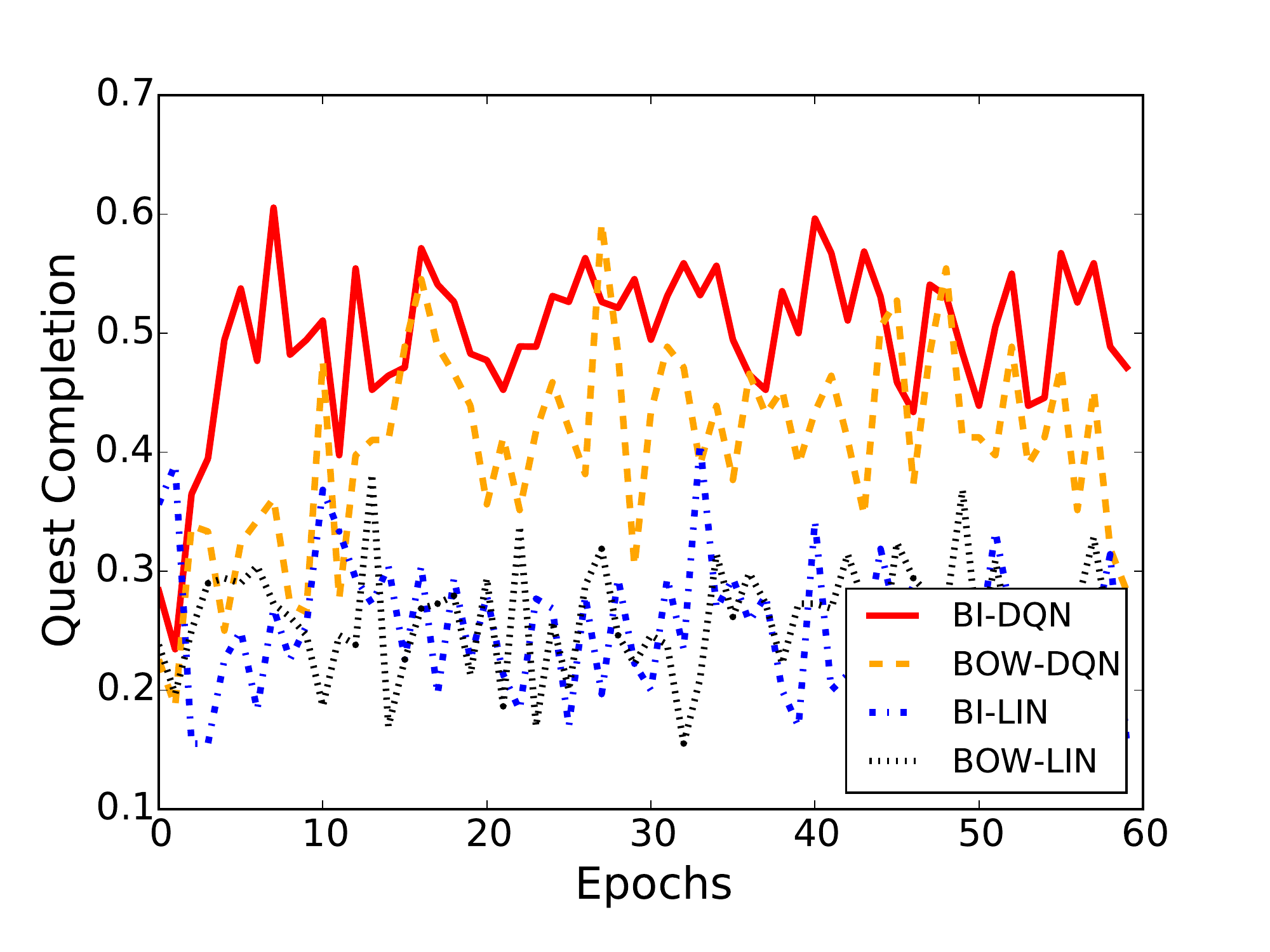}
\caption{Quest completion rates of DQN vs. Linear models on Home world.}
	\label{fig:dqn-simple}
\end{figure}

In addition, we also investigated the impact of using a deep neural network for modeling the action scorer $\phi_A$. Figure~\ref{fig:dqn-simple} illustrates the performance of the BOW-DQN and BI-DQN models along with their simpler versions BOW-LIN and BI-LIN, which use a single linear layer for $\phi_A$. It can be seen that the DQN models clearly achieve better performance than their linear counterparts, which points to them modeling the control policy better.

%since it may take two steps to move to the correct room.} 
% We also observe that the LSTM-DQN converges to a policy with higher Q-value over the validation set of states.

\paragraph{Fantasy World}
We evaluate all the models on the Fantasy world in the same manner as before and report reward, quest completion rates and Q-values. The quest we evaluate on involves crossing the broken bridge (which takes a minimum of five steps), with the possibility of falling off at random  (a 5\% chance) when the player is on the bridge. The game has an additional quest of reaching a secret tomb.
 However, this is a complex quest that requires the player to memorize game events and perform high-level planning which are beyond the scope of this current work. Therefore, we focus only on the first quest.
 
% our agent was succesful in completing only the first quest of crossing the bridge safely.
% Hence, we only report quest completion rates for the first quest of crossing the bridge safely. 

From Figure~\ref{fig:results} (bottom), we can see that the Random baseline does poorly in terms of both average per-episode reward\footnote{Note that the rewards graph is in log scale.} and quest completion rates. BOW-DQN converges to a much higher average reward of $-12.68$ and achieves around  82\% quest completion. Again, the BOW-DQN is often confused by varying  (10 different)  descriptions of the portions of the bridge, which reflects in its erratic performance on the quest. \imp{The BI-DQN  performs very well on quest completion by finishing 97\% of quests. However, this model tends to find sub-optimal solutions and gets an average reward of $-26.68$, even worse than BOW-DQN. One reason for this is the negative rewards the agent obtains after falling off the bridge.} The LSTM-DQN model again performs best, achieving an average reward of $-11.33$ and completing 96\% of quests on average. Though this world does not contain descriptions adversarial to BOW-DQN or BI-DQN, the LSTM-DQN obtains higher average reward by completing the quest in fewer steps and showing more resilience to variations in the state descriptions.
 
 % Even though the max Q-values on the validation set is lower for LSTM-DQN,
% We also find that the quest completion graph of LSTM-DQN is more stable which demonstrates its resilience to varying text descriptions in different episodes. 
% \todo[]{take a look}

%\subsection{Analysis}

%\begin{table*}[t]
%\centering
%\resizebox{\textwidth}{!}{%
%\begin{tabular}{| c | c |} \hline
%\textbf{Description} & \textbf{Nearest neighbor} \\ \hline
%This area has a bed, desk and a dresser. & You have arrived in the bedroom. You can rest here. \\
%You have arrived in the kitchen. You can find food and drinks here.	& This living area has pizza, coke, and icecream. \\
%This space has a swing, flowers and trees. & You have arrived at the garden. You can exercise here \\
%You have entered the living room. You can watch TV here. & This room has two sofas, chairs and a chandelier. \\ \hline
%\end{tabular}
%}
%\caption{Examples of room descriptions from the Home world and their nearest neighbors according to their vector representations from the LSTM representation generator.}
%\label{table:lstm-desc-examples}
%\end{table*}

\begin{table*}[t]
\centering
\resizebox{\textwidth}{!}{%
\begin{tabular}{|p {0.5\textwidth}|p{0.5\textwidth}|} \hline
\textbf{Description} & \textbf{Nearest neighbor} \\ \hline
{\footnotesize You are halfways out on the unstable bridge. From the castle you hear a distant howling sound, like that of a large dog or other beast.} & {\footnotesize The bridge slopes precariously where it extends westwards towards the lowest point - the center point of the hang bridge. You clasp the ropes firmly as the bridge sways and creaks under you.} \\ \hline
%{\footnotesize The bridge slopes precariously where it extends westwards towards the lowest point - the center point of the hang bridge. The section of rope you hold onto crumble in your hands, parts of it breaking apart. You sway trying to regain balance.} & {\footnotesize The bridge slopes precariously where it extends westwards towards the lowest point - the center point of the hang bridge. The bridge creaks under your feet. Those planks do not seem very sturdy.} \\ \hline
{\footnotesize The ruins opens up to the sky in a small open area, lined by columns. ... To the west is the gatehouse and entrance to the castle, whereas southwards the columns make way for a wide open courtyard.}	& {\footnotesize  The old gatehouse is near collapse. .... East the gatehouse leads out to a small open area surrounded by the remains of the castle.  There is also a standing archway offering passage to a path along the old southern inner wall.} \\ \hline
\end{tabular}
}
\caption{Sample descriptions from the Fantasy world and their nearest neighbors (NN) according to their vector representations from the LSTM representation generator. The NNs are often descriptions of the same or similar (nearby) states in the game.}
\label{table:lstm-desc-examples}
\end{table*}

\paragraph{Transfer Learning}
We would like the representations learnt by $\phi_R$ to be generic enough and \emph{transferable} to new game worlds. To test this, we created a second Home world with the same rooms, but a completely different map, changing the locations of the rooms and the pathways between them. The main differentiating factor of this world from the original home world lies in the high-level planning required to complete quests.

%\begin{figure}[!t]
%\centering
%\resizebox{0.9\columnwidth}{!}{%
%\includegraphics[]{fig/home-transfer-avgR}
%}
%\caption{Evolution of average Reward with/without transfer of learnt LSTM representations on Home world}
%\label{fig:home-transfer-avgR}
%\end{figure}
%
%\begin{figure}[!t]
%\centering
%\resizebox{0.9\columnwidth}{!}{%
%\includegraphics[]{fig/home-priority-avgR}
%}
%\caption{Evolution of average Reward with uniform vs priority sampling (Home world)}
%\label{fig:home-priority-avgR}
%\end{figure}

\todo[inline]{change above table to tutorial world}

We initialized the LSTM part of an LSTM-DQN agent with parameters $\theta_R$ learnt from the original home world and trained it on the new world.\footnote{The parameters for the Action Scorer ($\theta_A$) are initialized randomly.} Figure \ref{fig:results} (top right) demonstrates that the agent with transferred parameters is able to learn quicker than an agent starting from scratch initialized with random parameters (\emph{No~Transfer}), reaching the optimal policy almost 20 epochs earlier. This indicates that these simulated worlds can be used to learn good representations for language that transfer across worlds. 

\begin{figure}[!t]
\centering
\resizebox{\columnwidth}{!}{%
\includegraphics[]{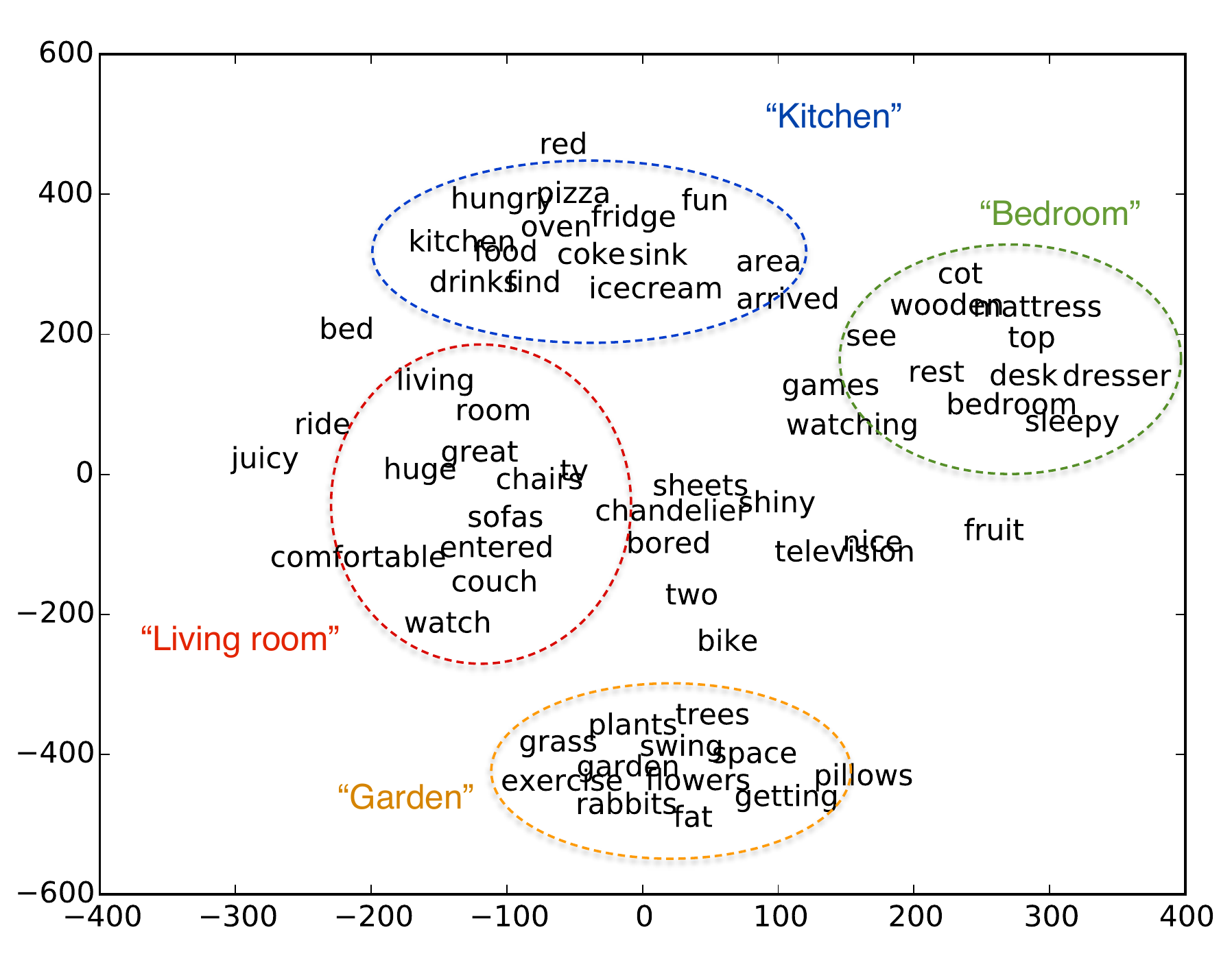}
}
\caption{t-SNE visualization of the word embeddings (except stopwords) after training on Home world. The embedding values are initialized randomly.}
\label{fig:tsne-lstm-random}
\end{figure}

\paragraph{Prioritized sampling}
We also investigate the effects of different minibatch sampling procedures on the parameter learning. From Figure~\ref{fig:results} (bottom right), we observe that using prioritized sampling significantly speeds up learning, with the agent achieving the optimal policy around 50 epochs faster than using uniform sampling. This shows promise for further research into different schemes of assigning priority to transitions.

\paragraph{Representation Analysis}
We analyzed the representations learnt by the LSTM-DQN model on the Home world. Figure~\ref{fig:tsne-lstm-random} shows a visualization of learnt word embeddings, reduced to two dimensions using t-SNE~\cite{van2008tsne}. All the vectors were initialized randomly before training. We can see that semantically similar words appear close together to form coherent subspaces. In fact, we observe four different subspaces, each for one type of room along with its corresponding object(s) and quest words. For instance, food items like \emph{pizza} and rooms like \emph{kitchen} are very close to the word \emph{hungry} which appears in a quest description. This shows that the agent learns to form meaningful associations between the semantics of the quest and the  environment.
Table~\ref{table:lstm-desc-examples} shows some examples of  descriptions from Fantasy world and their nearest neighbors using cosine similarity between their corresponding vector representations produced by LSTM-DQN.  The model is able to correlate descriptions of the same (or similar) underlying states and project them onto nearby points in the representation subspace.

\section{Conclusions}
We address the task of end-to-end learning of control policies for text-based games.  In these games, all interactions in the virtual world are through text and the underlying state is not observed. The resulting language variability makes such environments challenging for automatic game players. We employ a deep reinforcement learning framework to jointly learn state representations and action policies using game rewards as feedback. This framework enables us to map text descriptions into vector representations that capture the semantics of the game states.  \imp{Our experiments demonstrate the  importance of learning good representations of text in order to play these games well.} Future directions include tackling high-level planning and strategy learning to improve the performance of intelligent agents. 
% Code and data will be made available.
\section*{Acknowledgements}
\imp{We are grateful to the developers of Evennia, the game framework upon which this work is based. We also thank Nate Kushman, Clement Gehring, Gustavo Goretkin, members of MIT's NLP group and the anonymous EMNLP reviewers for insightful comments and feedback. T. Kulkarni was graciously supported by the Leventhal Fellowship. We would also like to acknowledge MIT's Center for Brains, Minds and Machines (CBMM) for support.}

% \bibliography{references}
% \bibliographystyle{acl2015}

\bibliographystyle{acl}
\bibliography{references}

\begin{thebibliography}{}

\bibitem[\protect\citename{Amato and Shani}2010]{amato2010high}
Christopher Amato and Guy Shani.
\newblock 2010.
\newblock High-level reinforcement learning in strategy games.
\newblock In {\em Proceedings of the 9th International Conference on Autonomous
  Agents and Multiagent Systems: Volume 1}, pages 75--82. International
  Foundation for Autonomous Agents and Multiagent Systems.

\bibitem[\protect\citename{Amir and Doyle}2002]{amir2002adventure}
Eyal Amir and Patrick Doyle.
\newblock 2002.
\newblock Adventure games: A challenge for cognitive robotics.
\newblock In {\em Proc. Int. Cognitive Robotics Workshop}, pages 148--155.

\bibitem[\protect\citename{Andreas and Klein}2015]{Andreas15Instructions}
Jacob Andreas and Dan Klein.
\newblock 2015.
\newblock Alignment-based compositional semantics for instruction following.
\newblock In {\em Proceedings of the Conference on Empirical Methods in Natural
  Language Processing}.

\bibitem[\protect\citename{Artzi and Zettlemoyer}2013]{artzi2013weakly}
Yoav Artzi and Luke Zettlemoyer.
\newblock 2013.
\newblock Weakly supervised learning of semantic parsers for mapping
  instructions to actions.
\newblock {\em Transactions of the Association for Computational Linguistics},
  1(1):49--62.

\bibitem[\protect\citename{Branavan \bgroup et al.\egroup
  }2010]{branavan2010reading}
SRK Branavan, Luke~S Zettlemoyer, and Regina Barzilay.
\newblock 2010.
\newblock Reading between the lines: Learning to map high-level instructions to
  commands.
\newblock In {\em Proceedings of the 48th Annual Meeting of the Association for
  Computational Linguistics}, pages 1268--1277. Association for Computational
  Linguistics.

\bibitem[\protect\citename{Branavan \bgroup et al.\egroup
  }2011a]{branavan2011learning}
SRK Branavan, David Silver, and Regina Barzilay.
\newblock 2011a.
\newblock Learning to win by reading manuals in a monte-carlo framework.
\newblock In {\em Proceedings of the 49th Annual Meeting of the Association for
  Computational Linguistics: Human Language Technologies-Volume 1}, pages
  268--277. Association for Computational Linguistics.

\bibitem[\protect\citename{Branavan \bgroup et al.\egroup
  }2011b]{branavan2011nonlinear}
SRK Branavan, David Silver, and Regina Barzilay.
\newblock 2011b.
\newblock Non-linear monte-carlo search in {C}ivilization {II}.
\newblock AAAI Press/International Joint Conferences on Artificial
  Intelligence.

\bibitem[\protect\citename{Curtis}1992]{curtis1992mudding}
Pavel Curtis.
\newblock 1992.
\newblock Mudding: Social phenomena in text-based virtual realities.
\newblock {\em High noon on the electronic frontier: Conceptual issues in
  cyberspace}, pages 347--374.

\bibitem[\protect\citename{DePristo and Zubek}2001]{depristo2001being}
Mark~A DePristo and Robert Zubek.
\newblock 2001.
\newblock being-in-the-world.
\newblock In {\em Proceedings of the 2001 AAAI Spring Symposium on Artificial
  Intelligence and Interactive Entertainment}, pages 31--34.

\bibitem[\protect\citename{Eisenstein \bgroup et al.\egroup
  }2009]{eisenstein-EtAl:2009:EMNLP}
Jacob Eisenstein, James Clarke, Dan Goldwasser, and Dan Roth.
\newblock 2009.
\newblock Reading to learn: Constructing features from semantic abstracts.
\newblock In {\em Proceedings of the Conference on Empirical Methods in Natural
  Language Processing}, pages 958--967, Singapore, August. Association for
  Computational Linguistics.

\bibitem[\protect\citename{Gorniak and Roy}2005]{DBLP:conf/aiide/GorniakR05}
Peter Gorniak and Deb Roy.
\newblock 2005.
\newblock Speaking with your sidekick: Understanding situated speech in
  computer role playing games.
\newblock In R.~Michael Young and John~E. Laird, editors, {\em Proceedings of
  the First Artificial Intelligence and Interactive Digital Entertainment
  Conference, June 1-5, 2005, Marina del Rey, California, {USA}}, pages 57--62.
  {AAAI} Press.

\bibitem[\protect\citename{Hochreiter and Schmidhuber}1997]{hochreiter1997long}
Sepp Hochreiter and J{\"u}rgen Schmidhuber.
\newblock 1997.
\newblock Long short-term memory.
\newblock {\em Neural computation}, 9(8):1735--1780.

\bibitem[\protect\citename{Kollar \bgroup et al.\egroup
  }2010]{kollar2010toward}
Thomas Kollar, Stefanie Tellex, Deb Roy, and Nicholas Roy.
\newblock 2010.
\newblock Toward understanding natural language directions.
\newblock In {\em Human-Robot Interaction (HRI), 2010 5th ACM/IEEE
  International Conference on}, pages 259--266. IEEE.

\bibitem[\protect\citename{Koutn{\'\i}k \bgroup et al.\egroup
  }2013]{koutnik2013evolving}
Jan Koutn{\'\i}k, Giuseppe Cuccu, J{\"u}rgen Schmidhuber, and Faustino Gomez.
\newblock 2013.
\newblock Evolving large-scale neural networks for vision-based reinforcement
  learning.
\newblock In {\em Proceedings of the 15th annual conference on Genetic and
  evolutionary computation}, pages 1061--1068. ACM.

\bibitem[\protect\citename{Kushman \bgroup et al.\egroup
  }2014]{kushman2014learning}
Nate Kushman, Yoav Artzi, Luke Zettlemoyer, and Regina Barzilay.
\newblock 2014.
\newblock Learning to automatically solve algebra word problems.
\newblock {\em ACL (1)}, pages 271--281.

\bibitem[\protect\citename{Matuszek \bgroup et al.\egroup
  }2013]{matuszek2013learning}
Cynthia Matuszek, Evan Herbst, Luke Zettlemoyer, and Dieter Fox.
\newblock 2013.
\newblock Learning to parse natural language commands to a robot control
  system.
\newblock In {\em Experimental Robotics}, pages 403--415. Springer.

\bibitem[\protect\citename{Mikolov \bgroup et al.\egroup
  }2013]{mikolov2013efficient}
Tomas Mikolov, Kai Chen, Greg Corrado, and Jeffrey Dean.
\newblock 2013.
\newblock Efficient estimation of word representations in vector space.
\newblock {\em arXiv preprint arXiv:1301.3781}.

\bibitem[\protect\citename{Mnih \bgroup et al.\egroup }2015]{mnih2015dqn}
Volodymyr Mnih, Koray Kavukcuoglu, David Silver, Andrei~A. Rusu, Joel Veness,
  Marc~G. Bellemare, Alex Graves, Martin Riedmiller, Andreas~K. Fidjeland,
  Georg Ostrovski, Stig Petersen, Charles Beattie, Amir Sadik, Ioannis
  Antonoglou, Helen King, Dharshan Kumaran, Daan Wierstra, Shane Legg, and
  Demis Hassabis.
\newblock 2015.
\newblock Human-level control through deep reinforcement learning.
\newblock {\em Nature}, 518(7540):529--533, 02.

\bibitem[\protect\citename{Moore and Atkeson}1993]{moore1993prioritized}
Andrew~W Moore and Christopher~G Atkeson.
\newblock 1993.
\newblock Prioritized sweeping: Reinforcement learning with less data and less
  time.
\newblock {\em Machine Learning}, 13(1):103--130.

\bibitem[\protect\citename{Pennington \bgroup et al.\egroup
  }2014]{pennington2014glove}
Jeffrey Pennington, Richard Socher, and Christopher~D Manning.
\newblock 2014.
\newblock Glove: Global vectors for word representation.
\newblock {\em Proceedings of the Empiricial Methods in Natural Language
  Processing (EMNLP 2014)}, 12.

\bibitem[\protect\citename{Silver \bgroup et al.\egroup
  }2007]{silver2007reinforcement}
David Silver, Richard~S Sutton, and Martin M{\"u}ller.
\newblock 2007.
\newblock Reinforcement learning of local shape in the game of go.
\newblock In {\em IJCAI}, volume~7, pages 1053--1058.

\bibitem[\protect\citename{Sutskever \bgroup et al.\egroup
  }2014]{sutskever2014sequence}
Ilya Sutskever, Oriol Vinyals, and Quoc~VV Le.
\newblock 2014.
\newblock Sequence to sequence learning with neural networks.
\newblock In {\em Advances in Neural Information Processing Systems}, pages
  3104--3112.

\bibitem[\protect\citename{Sutton and Barto}1998]{sutton1998introduction}
Richard~S Sutton and Andrew~G Barto.
\newblock 1998.
\newblock {\em Introduction to reinforcement learning}.
\newblock MIT Press.

\bibitem[\protect\citename{Szita}2012]{szita2012reinforcement}
Istv{\'a}n Szita.
\newblock 2012.
\newblock Reinforcement learning in games.
\newblock In {\em Reinforcement Learning}, pages 539--577. Springer.

\bibitem[\protect\citename{Tai \bgroup et al.\egroup }2015]{tai2015improved}
Kai~Sheng Tai, Richard Socher, and Christopher~D. Manning.
\newblock 2015.
\newblock Improved semantic representations from tree-structured long
  short-term memory networks.
\newblock In {\em Proceedings of the 53rd Annual Meeting of the Association for
  Computational Linguistics and the 7th International Joint Conference on
  Natural Language Processing (Volume 1: Long Papers)}, pages 1556--1566,
  Beijing, China, July. Association for Computational Linguistics.

\bibitem[\protect\citename{Tieleman and Hinton}2012]{tieleman2012lecture}
Tijmen Tieleman and Geoffrey Hinton.
\newblock 2012.
\newblock Lecture 6.5-rmsprop: Divide the gradient by a running average of its
  recent magnitude.
\newblock {\em COURSERA: Neural Networks for Machine Learning}, 4.

\bibitem[\protect\citename{Van~der Maaten and Hinton}2008]{van2008tsne}
Laurens Van~der Maaten and Geoffrey Hinton.
\newblock 2008.
\newblock Visualizing data using t-sne.
\newblock {\em Journal of Machine Learning Research}, 9(2579-2605):85.

\bibitem[\protect\citename{Vogel and Jurafsky}2010]{vogel2010learning}
Adam Vogel and Dan Jurafsky.
\newblock 2010.
\newblock Learning to follow navigational directions.
\newblock In {\em Proceedings of the 48th Annual Meeting of the Association for
  Computational Linguistics}, pages 806--814. Association for Computational
  Linguistics.

\bibitem[\protect\citename{Watkins and Dayan}1992]{watkins1992q}
Christopher~JCH Watkins and Peter Dayan.
\newblock 1992.
\newblock Q-learning.
\newblock {\em Machine learning}, 8(3-4):279--292.

\end{thebibliography}

\end{document}